\documentclass[runningheads]{llncs}
\usepackage{graphicx}
\usepackage[square,comma,numbers,sort&compress]{natbib}
\setcitestyle{numbers,square,citesep={,},aysep={,},yysep={;}}
\makeatletter
\renewcommand{\@biblabel}[1]{#1.}
\makeatother

\usepackage{gensymb}
\usepackage{footmisc}
\usepackage{amsmath}
\usepackage{amssymb}
\usepackage{mathtools}
\usepackage[retainorgcmds]{IEEEtrantools}

\usepackage{caption}
\usepackage[font=scriptsize, labelfont=scriptsize, textfont=scriptsize]{subfig}
\usepackage{booktabs}
\usepackage{array}
\newcommand{\tilda}{\raise.22ex\hbox{$\scriptstyle\sim$}}
\usepackage{multirow}

\usepackage{xcolor}
\usepackage{hyperref}

\usepackage[misc,geometry]{ifsym}

\begin{document}
\title{Full Page Handwriting Recognition via Image to Sequence Extraction}
\titlerunning{Full Page HTR via Image to Seq Extraction (Published at ICDAR 2021)}
\author{Sumeet S. Singh\inst{1}(\Letter)\orcidID{0000-0002-5323-9678} \and
Sergey Karayev\inst{1}
}
\authorrunning{S. S. Singh et al.}
\institute{\textsuperscript{1} Turnitin, 2101 Webster St \#1800, Oakland, CA 94612, USA\\
\email{ssingh@turnitin.com} / \email{sumeet@singhonline.info}, \email{sergeykarayev@gmail.com}
}

\maketitle              % typeset the header of the contribution
\begin{abstract}
We present a Neural Network based Handwritten Text Recognition (HTR) model architecture that can be trained to recognize full pages of handwritten or printed text without image segmentation. Being based on Image to Sequence architecture, it can extract text present in an image and then sequence it correctly without imposing any constraints regarding orientation, layout and size of text and non-text. Further, it can also be trained to generate auxiliary markup related to formatting, layout and content. We use character level vocabulary, thereby enabling language and terminology of any subject. The model achieves a new state-of-art in paragraph level recognition on the IAM dataset. When evaluated on scans of real world handwritten free form test answers - beset with curved and slanted lines, drawings, tables, math, chemistry and other symbols - it performs better than all commercially available HTR cloud APIs. It is deployed in production as part of a commercial web application.

% \keywords{Handwriting Recognition \and Neural Networks}
\end{abstract}

\section{Overview}
With the advancement of Deep Neural Networks, the field of {\small HTR} has progressed by leaps and bounds.
Neural Networks have enabled algorithm developers to increasingly rely on features and algorithms learned from data rather than hand crafted ones.
This makes it easier than ever before to create {\small HTR} models for new datasets and languages.
That said, {\small HTR} models still embody domain-specific inductive biases, assumptions and heuristics in their architectures regarding the layout, size and orientation of text.
We aim to address some of these limitations in this work.

\subsection{Challenges of Real World Full Page HTR}
\label{sec-free-form-dataset}

\begin{figure}%[ht]
    \centering
    % \subfloat[Indented computer source code.]{
    %     \includegraphics[width=0.3\textwidth,keepaspectratio]{Figures/Fullpage_code_41.jpg}
    %     \label{fig-fullpage-code}
    % }
    \subfloat[]{
        \includegraphics[width=0.3405\textwidth]{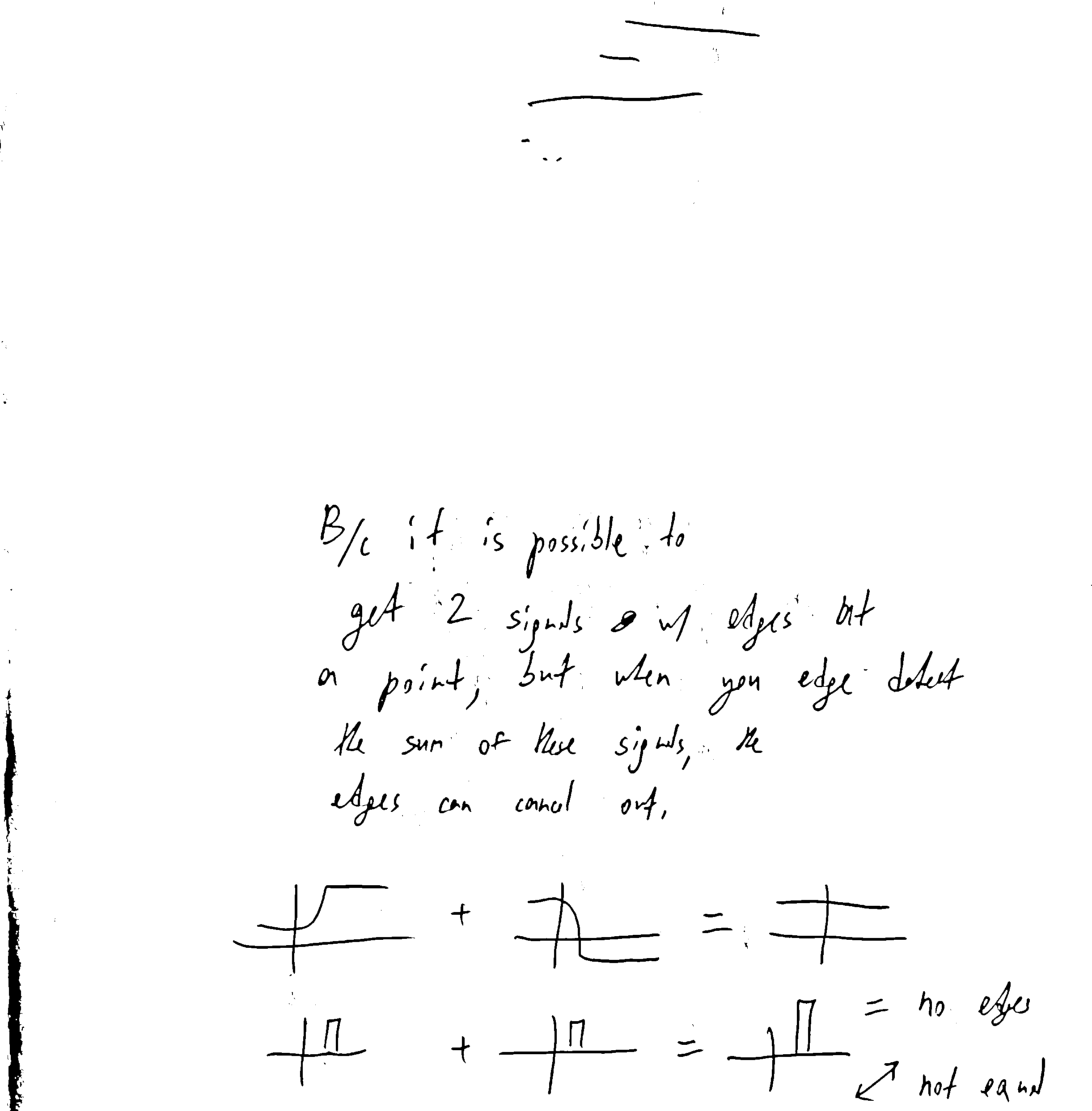}
        \label{fig-fullpage-answer1}
    }
    % \hspace{10pt}
    \vrule
    % \hspace{13pt}
    \subfloat[]{
        \includegraphics[width=0.35\textwidth]{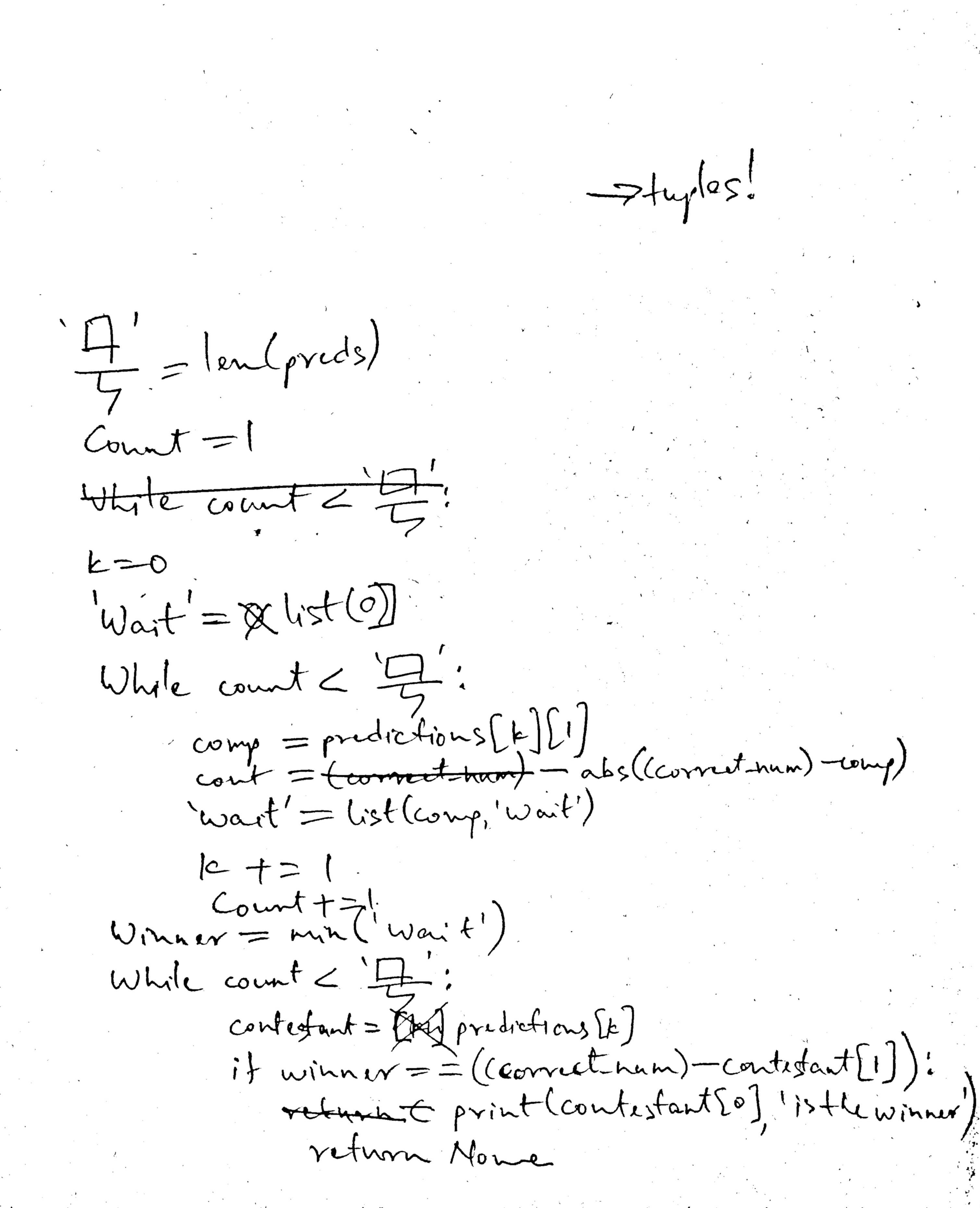}
        \label{fig-fullpage-code}
    }
    \\
    % \vspace{4pt}
    \hrule
    % \vspace{4pt}  
    \subfloat[]{
        \includegraphics[width=0.45\textwidth]{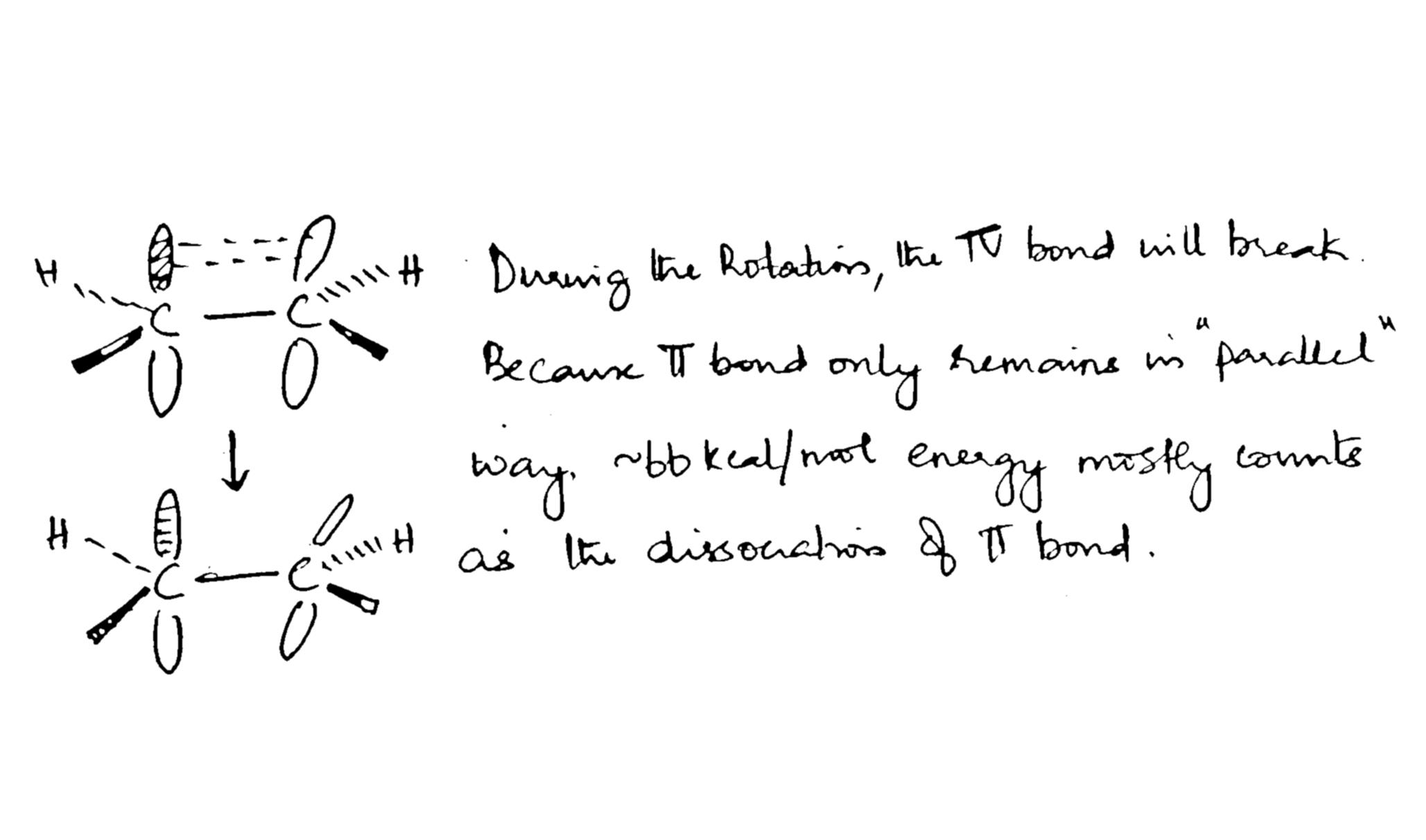}
        \label{fig-chemistry}
    }
    \hfill
    \vrule
    \hfill
    \subfloat[]{
        \includegraphics[width=0.45\textwidth]{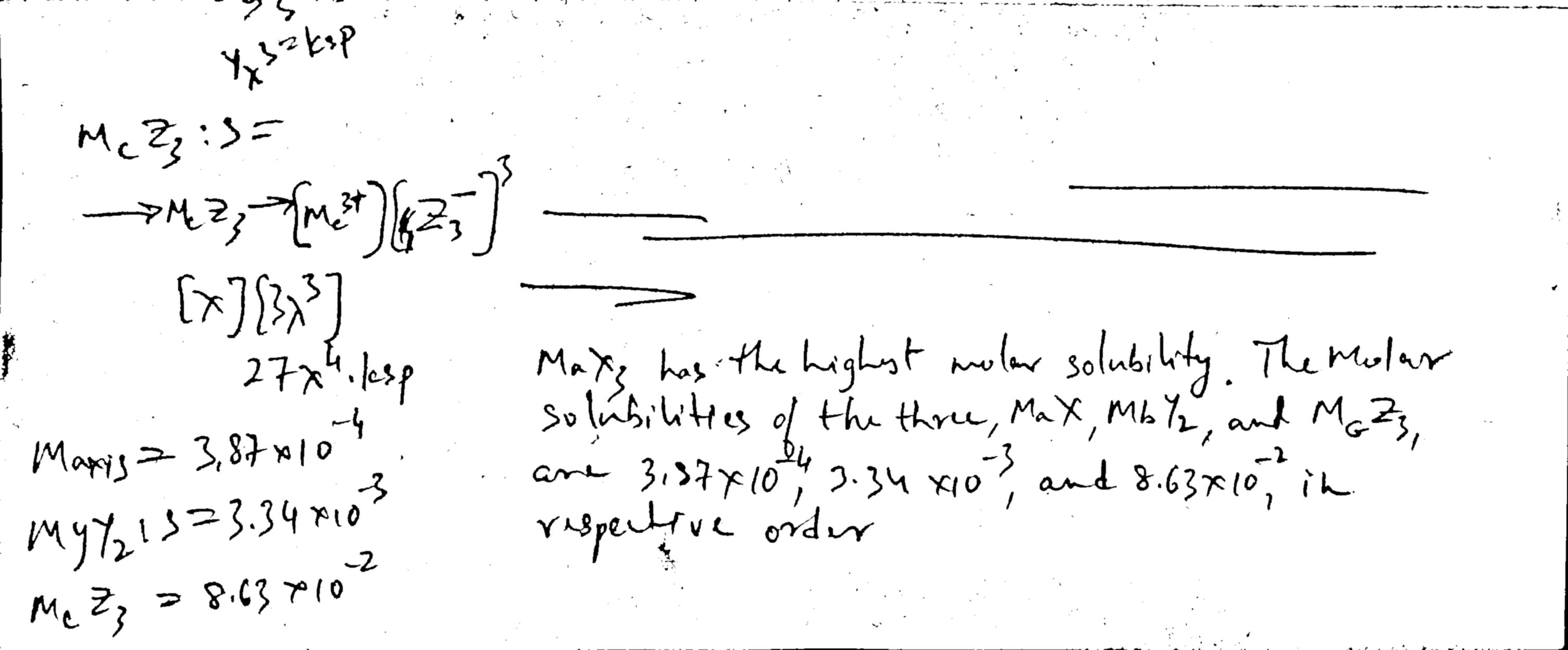}
        \label{fig-math}
    }\\
        % \bottomrule
    \caption{Samples from our Free Form Answers dataset. (a) Full page text with drawing. (b) Full page computer source code. (c) Diagrams and text with embedded math. (d) Math and text regions, embedded math and stray artifacts.}
    \label{fig-answers1}
\end{figure}

\begin{figure}%[ht]
    \centering
    \begin{tabular}{c|c}
        \multirow{2}{*}[1.7cm]{
            \subfloat[Test image]{
                \includegraphics[width=.45\textwidth]{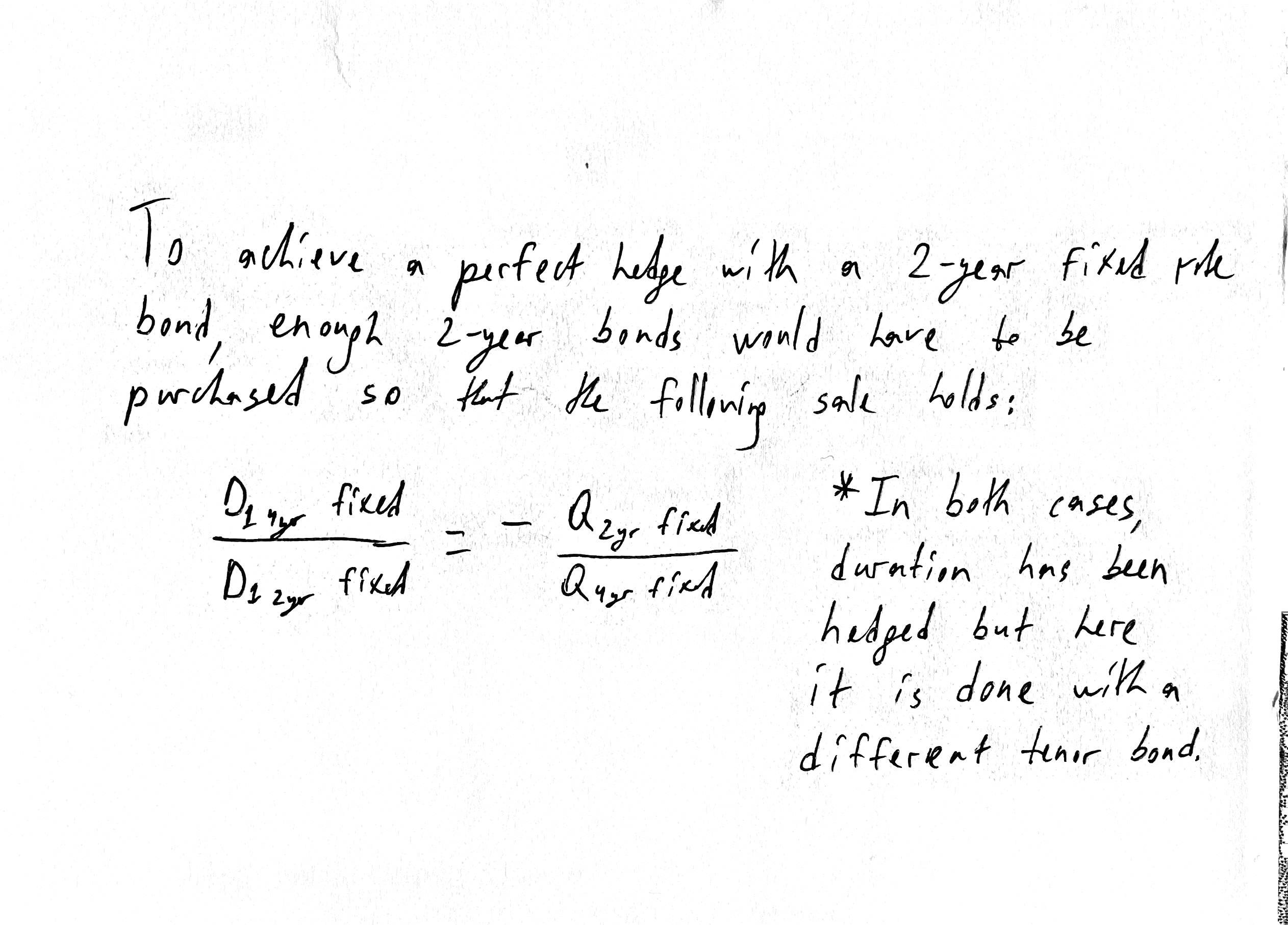}
            }
        } & \subfloat[Target label]{
            \includegraphics[width=.45\textwidth]{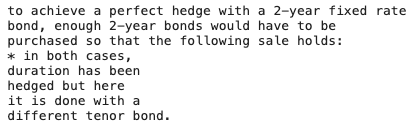}
        }
        \\
\        & \subfloat[Model's prediction]{
            \includegraphics[width=.45\textwidth]{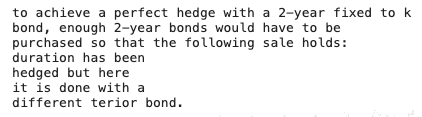}
            \label{fig-pred-4717}
        }
    \end{tabular}
    \caption{Test image with two text and one math regions. The model successfully skips the math region and transcribes the two text regions, with some mistakes.}
    \label{fig-answers2}
\end{figure}

Our target data - \textit{Free Form Answers} dataset - was derived from scans of test paper submissions of STEM subjects, from school level all the way up to post-graduate courses. It consists of images containing possibly multiple blurbs of handwritten text, math equations, tables, drawings, diagrams, side-notes, scratched out text and text inserted using an arrow / circumflex and other artifacts, all put together with no reliable layout (Figures~\ref{fig-answers1} and \ref{fig-answers2}). Length of transcription ranges from 0 to 1100 characters averaging around 160.

\subsubsection{Problem Framework}
\label{sec-problem-framework}
In order to get a handle on this seemingly random structure, we define each independent blurb as a \textit{region} and view each image as being composed of one or more regions. The regions are further classified as text, math, drawing, tables and deleted (scratched) text. Non textual regions and deleted text (hereafter \textit{untranscribed regions} ) are optionally demarcated with special \textit{auxiliary tags} but left untranscribed otherwise. Text regions range from a few characters to multiple paragraphs and are possibly interspersed or embedded with untranscribed regions. Text was transcribed in the order it was meant to be read\footnote{This becomes relevant when text is not horizontal or when inserted using a circumflex or arrow.} and line ends, empty lines, spaces and indentations are also faithfully transcribed. These can be programmatically removed later if so desired. Since the regions can be randomly  situated on the image, we define a predictable region sequencing order \textit{the natural reading order} as the order in which somebody would read the answer aloud. This order is implicitly captured in the transcription and the model must learn to reproduce it.

% \subsubsection{Problem Definition}
With the above framework in mind, the problem becomes: 1) Implicitly identify and classify regions on the image 2) Extract text from each text region and emit it in natural reading order 3) Ignore / skip unrecognized artifacts and 4) Produce auxiliary markup. We call the above formulation \emph{extractive}, since it seeks to extract the identified and desirable content but ignore the unknown and undesirable. This formulation is generic enough to cover simple to complex scenarios but also teaches the model to skip over artifacts that were not encountered in the dataset.

\subsection{Limitations of Existing Architectures}
\label{sec-limitations-of-SOTA}
Most state-of-the-art HTR models rely on a prior image segmentation method to cleanly isolate pieces of text (words, lines or paragraphs).
There are several problems with this approach.

First, image segmentation in such models is a separate system, usually based on hand-crafted features and heuristics which do not hold up when the data changes significantly.
Some methods go even further and correct segmented text for slant, curve, and height, using the same problematic features and heuristics.

Second, and more importantly, clean segmentation of units of text is not even possible in many cases of real world handwriting.
For instance, observe that lines are curved or interspersed with non-textual symbols and artifacts in Figures \ref{fig-answers1} and \ref{fig-answers2}.
Further discussion of such limitations can be found in the literature \citep{Bluche2016JointLS,DBLP:journals/corr/BlucheLM16}.

Third, stitching a complete transcription from the individually transcribed text regions introduces yet another system, with its own potentials for errors, and brittleness to changing the data distribution (for example, right-to-left languages would require a different stitching system).

Lastly, formatting and indentation tends to get lost in this three-step process because text segments are usually concatenated using space or newline as a separator.
This would be unacceptable when transcribing Python language source code, for instance.

In our end-to-end model all of the above steps are implicit and learned from data.
Adapting the model to a new dataset or adding new capabilities is just a matter of retraining or fine-tuning with different labels or different data.

\subsection{Model Overview}

\begin{figure}
\centering
    \subfloat[][Model Outline]{
        \includegraphics[width=0.62\textwidth]{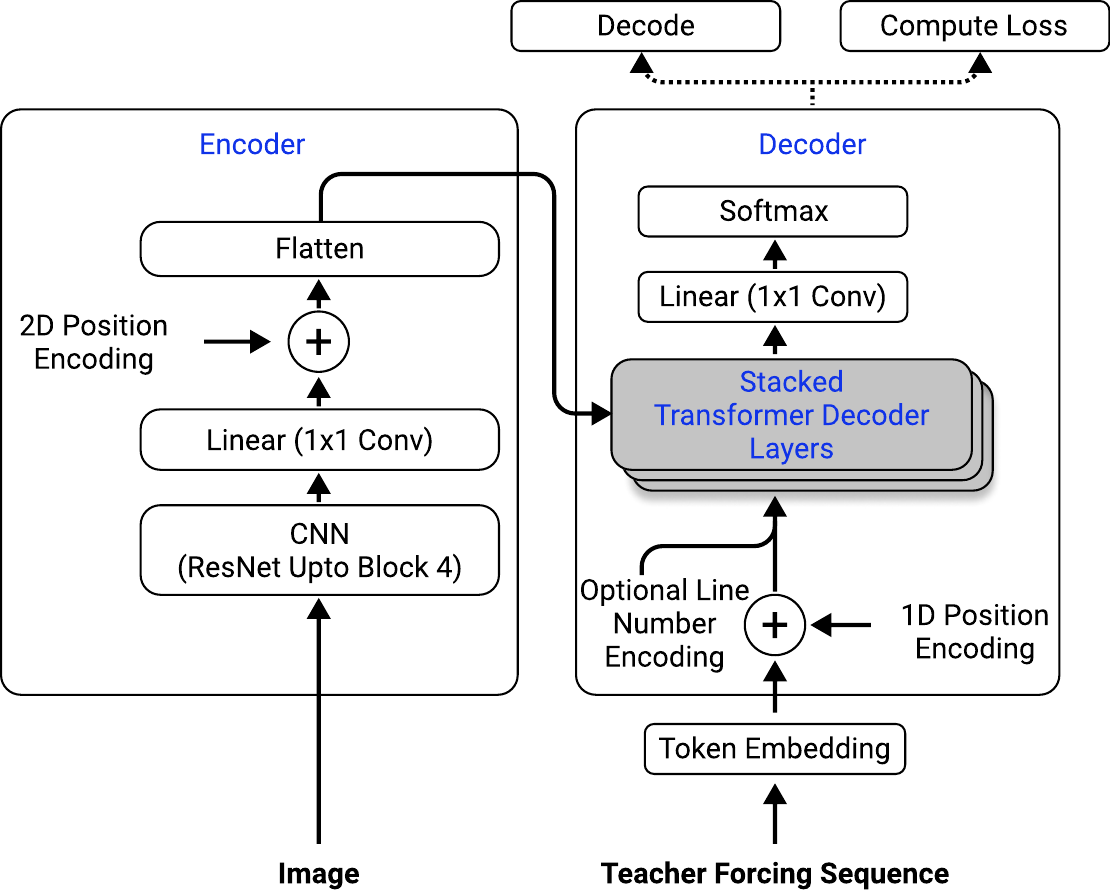}
        \label{fig-model-outline}
    }
    \hfill
    \subfloat[][Decoder Transformer Layer]{
        \includegraphics[width=0.33\textwidth]{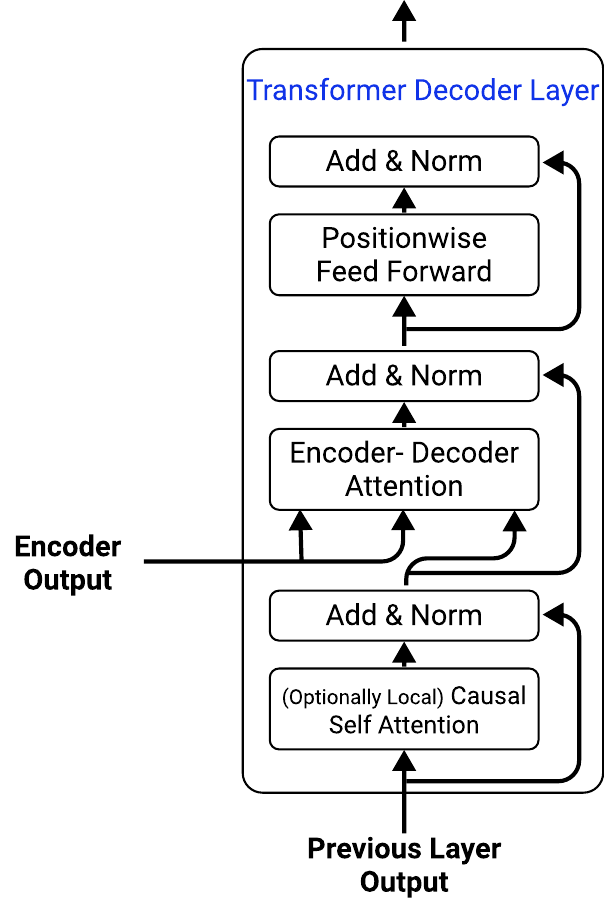}
        \label{fig-decoder-layer}
    }
\caption{\small Model Schematics. Left: CNN Encoder and Transformer Decoder. Right: The Transformer Layer is identical to \citep{DBLP:journals/corr/VaswaniSPUJGKP17} except that Self Attention maybe optionally Local.
         Teacher Forcing Sequence refers to the ground-truth shifted right during training or the predicted tokens when predicting.}
\label{fig-model}
\end{figure}

\subsubsection{Image to Sequence Architecture}
Our model (\autoref{fig-model}) belongs to the lineage of attention-based sequence-to-sequence and tensor-to-tensor architectures \citep{Xu2015ShowAA,DBLP:journals/corr/abs-1803-07416}.
It consists of a ResNet \citep{DBLP:journals/corr/HeZRS15} \emph{encoder} and a Transformer \citep{DBLP:journals/corr/VaswaniSPUJGKP17} \emph{decoder}. It has a \emph{Image to Sequence architecture} i.e., it learns to map an image to a sequence of tokens.
The model signals end of output sequence via a special \textsc{\textless{eos\textgreater}} token, so there is no limit on sequence length\footnote{Except a limit set at prediction to prevent an endless loop.}.

\subsubsection{Formatting and Auxiliary Markup}
We trained the model to extract text from text regions only, and to either skip the the untranscribed regions or produce special markup tags (\textsc{ \textless{end-of-region\textgreater}, \textless{math\textgreater}, \textless{deleted-text\textgreater}, \textless{table\textgreater} and \textless{drawing\textgreater}}) for them.
This requires the model to identify the untranscribed regions, which are almost always \emph{unique per sample}.
It generalizes reasonably well in either of these tasks.
Similarly, because whitespace is preserved in our ground truth transcription, the model learns  to faithfully replicate multiple empty lines and indentation spaces, which is important for computer source code for example.

\subsubsection{Token Vocabulary}
In order to cater to a variety of subject matter terminology and proper nouns, we chose to use character level token vocabularies in our base configuration.
That said, we experimented with different types of vocabularies, for example character vs hybrid word/character, or lowercase vs mixed case.
Model performance did not differ significantly between these variations.
Therefore, we are confident that more complex transcription tasks -- such as transcribing math \citep{Deng2017ImagetoMarkupGW,DBLP:journals/corr/abs-1802-05415} and complex layouts such as tables -- should be possible given appropriately transcribed training data and a character-level vocabulary.

\subsubsection{Layout Agnostic Architecture}
Since there are no assumptions regarding the layout of text, the model works equally well on single-column vs two-column text, and easily learns to produce a column separator tag (\autoref{tab-free-response}).
The multistage HTR models described in \autoref{sec-limitations-of-SOTA} on the other hand, would concatenate adjacent lines from the two columns, unless a separate code module was added to recognize columns and modifications made to the rest of the code.
% Furthermore, in one experiment we randomly rotated the samples in the range -40 to +40 degrees with no issues in performance.

\subsubsection{Performance}
The model outperforms the best commercially available API on our proprietary Free Form Answers dataset (\autoref{tab-free-response}) and state of art full page recognition \citep{DBLP:journals/corr/BlucheLM16} and full paragraph recognition \citep{Bluche2016JointLS} models on the academic IAM dataset (\autoref{tab-IAM}).

\section{Related Work}
The first work to use Neural Networks for offline {\small HTR} was by \citet{726791} which recognized single digits. The next evolution was by \citet{Graves2008OfflineHR} who used an {\small MDLSTM} model and {\small CTC} loss and decoding \citep{Graves2006ConnectionistTC} to recognize Tunisian town names in Arabic. Their model was designed to recognize a single line of text, oriented horizontally. %\footnote{The vertical dimension of the 2D feature map was collapsed, resulting in a 1D feature sequence.}.
\cite{DBLP:journals/corr/PhamKL13} refined the MDLSTM+CTC model by adding dropout, but it remained a single-line recognition model dependent on prior line segmentation and normalization.
\citet{7814068} further increased performance by adding deslanting preprocessing (which is entirely dataset specific) and using a bigger network.

\citet{DBLP:journals/corr/BlucheLM16} developed a model with the same vision as ours: full-page recognition with no layout or size assumptions.
However, they subsequently abandoned this approach citing prohibitive memory requirements, unavailability of GPU acceleration for training of {\small MDLSTM} and intractable inference time.
% \footnote{
    % The issues stem primarily from the model's heavy use of {\small MDLSTM} and {\small LSTM} which inherently require sequential input and therefore cannot be parallelized.
    % Additionally, the {\small MDLSTM} used to compute step-wise attention in that model had to scan the entire image four times for each character predicted.
% }.
Follow-up work by the same authors \citep{Bluche2016JointLS,8270042} saw a come-back to the encoder-only + {\small CTC} approach but with a scaled-back {\small MDLSTM} attention model that could isolate individual lines of text, in effect performing automatic line segmentation and enabling the model to recognize paragraphs.
While far superior to single-line recognition models, this approach still hard-codes the assumption that lines of text stretch horizontally from left to right, and fill the entire image.
Therefore, it can't handle arbitrary text layouts and instead, relies on paragraph segmentation.
This approach also does not output a variable-length sequence rather a fixed number of lines $T$ (some of which may be empty), $T$ being baked into the model during training.
Finally, since the predicted lines are joined together using a fixed separator it is unlikely that the model can faithfully reproduce empty lines and indentation.

Further on, embracing the trend towards more parallel architectures, \citet{8270042} and \citet{8269951} replaced the {\small MDLSTM} encoder with {\small CNN}, but continued to rely on {\small CTC}.
% In its ongoing work to make models faster, more efficient, and capture longer range dependencies, the deep learning community has been steadily moving away from sequential architectures ({\small MDLSTM / LSTM / RNN}) and towards parallel architectures such as {\small CNN} and the Transformer. Consistent with this trend, 

Another recent trend in deep learning has been the move away from encoder-only towards encoder-decoder sequence-to-sequence architectures. These architectures decouple the sequence length of the output from the input %\footnote{This is done by enabling the decoder to emit a special end-of-sequence token.}
thereby eliminating the need for {\small CTC} loss and decoding. Cross-entropy loss and greedy / beam-search decoding which are less compute-intensive are employed instead. \citet{8978049} applied such architecture to recognizing very short passages of japanese text (a handful of short vertical lines). Similarly \citet{wang2019decoupled} applied a bespoke sequence-to-sequence architecture for recognizing single lines.

It is well established now that Transformers can completely replace {\small LSTMs} in function and are more parameter efficient, accurate and enable longer sequences.
Embracing this trend \citet{kang2020pay} published the most `modern' architecture to date employing {\small CNN} and Transformers somewhat like ours. However since it collapses the vertical dimension of the image feature map, this model is designed to recognize single lines only. It also employs a transformer encoder thereby making it larger than our model. To our knowledge, ours is the only work other than \citep{DBLP:journals/corr/BlucheLM16} that attempts full page handwriting recognition.

\section{Model Architecture}
Our Neural Network architecture is shown in \autoref{fig-model}.
It is an encoder-decoder architecture, using ResNet \citep{DBLP:journals/corr/HeZRS15} for encoding the image, and Transformer \citep{DBLP:journals/corr/VaswaniSPUJGKP17} for decoding the encoded representation into text.
We refer you to \citep{Xu2015ShowAA,DBLP:journals/corr/abs-1802-05415,DBLP:journals/corr/VaswaniSPUJGKP17,DBLP:journals/corr/SutskeverVL14} for a background on neural image-to-sequence and sequence-to-sequence models.
This section will fill in the remaining details necessary to reproduce our model architecture.

We use the term \emph{base configuration} to refer to our most frequently used model configuration, and all configuration parameters we list hereafter describe this configuration unless stated otherwise.

\subsubsection{Encoder}
 The encoder uses a {\small CNN} to extract a 2D feature-map from the input image.
 It uses the ResNet architecture without its last two layers: the average-pool and linear projection.
 The feature-map is then projected to match the Transformer's hidden-size $d_{model}$, then a 2D positional encoding added and finally flattened into a 1D sequence.
 2D positional encoding is a fixed sinusoidal encoding as in \citep{DBLP:journals/corr/VaswaniSPUJGKP17}, but using the first $d_{model}/2$ channels to encode the Y coordinate and the rest to encode the X coordinate (\autoref{eqn-pos-encoding}) (similar to \citep{parmar2018image}).
 Output $\boldsymbol{I}$ of the \emph{Flatten} layer is made available to all Transformer decoder layers, as is standard.

\begin{IEEEeqnarray*}{rCl}
    PE(y, 2i) &=& sin(y / 10000^{2i/d_{model}}) \\
    PE(y, 2i+1) &=& cos(y / 10000^{2i/d_{model}}) \\
    PE(x, d_{model}/2 + 2i) &=& sin(x / 10000^{2i/d_{model}}) \IEEEyesnumber \label{eqn-pos-encoding}\\
    PE(x, d_{model}/2 + 2i+1) &=& cos(x / 10000^{2i/d_{model}}) \\
    i & \in & [0, d_{model}/4)
\end{IEEEeqnarray*}

\subsubsection{Decoder}
The decoder is a Transformer stack with non-causual attention to the encoder output (its layers can attend to the encoder's entire output) and causal self-attention (it can only attend to past positions of its text input).
As is standard, training is done with \textit{teacher forcing}, which means that the ground truth text input is shifted one off from the output.

In total, the base configuration has 27.8 million parameters (6.3M decoder, 21.4M ResNet).

The input vectors are enhanced with 1D position encoding, as is standard.
Additionally, we concatenate a \textit{line number encoding} ($lne$) - the scaled text line number ($l$) that the token lies on - to it (Figure \ref{fig-model-outline}). Assuming a maximum of 100 text lines, $l \in [1,100]$ and $lne = l / 100$.
% Line number $l$ is the position of a line of text of the target sequence in \textit{natural reading order}, starting from 1 up to a max configurable value (100).
% Line number encoding $lne$ is obtained by scaling $l$ to fit within [0,1] i.e., $lne = l / 100$.
We added $lne$ to the model in order to address line level errors i.e., missing or duplicated lines but it was applied only in \autoref{tab-IAM}. We haven't yet officially concluded on its impact on model performance and mention it here only for completeness sake.

In order to improve memory and computation requirements of our model, we implemented a localized form of causal self-attention by limiting the attention span to 50 (configurable) past positions.
This is similar to Sliding Window Attention of \citep{beltagy2020longformer}) or a 1D version of Local Self Attention of \citep{parmar2018image}).
We hypothesized that a look back of 50 characters should be enough to satisfy the language modeling needs of our task, while the limited attention span should help training converge faster, both assumptions being validated by experiment.
Final model performance however, was not impacted by it. That said, a thorough ablation study was not performed.
Practically though, it allowed us to use larger mini-batches by about 12\%.

\subsubsection{Objective Function}
% The model output at each step (position) $t$ of the output sequence is viewed as a discrete random variable $\boldsymbol{Y}_t$ that takes its values from the vocabulary set $\lbrace 1, \ldots , V \rbrace$. The model outputs $\boldsymbol{Y}_t$'s discrete probability distribution $\boldsymbol{p}_t$ (\autoref{eqn-step-prob}) conditioned upon the tokens generated thus far $\boldsymbol{y}_{<t}$ and the input image $\boldsymbol{I}$ (\autoref{eqn-step-cond-prob}).
For each step $t$ the model outputs the token probability distribution $\boldsymbol{p}_t$ over the vocabulary set $\lbrace 1, \ldots , V \rbrace$ (\autoref{eqn-step-prob}).
This distribution is conditioned upon the tokens generated thus far $\boldsymbol{y}_{<t}$ and $\boldsymbol{I}$ (\autoref{eqn-step-cond-prob}).
Probability of the entire token sequence is therefore given by \autoref{eqn-sequence-prob}.
\begin{IEEEeqnarray}{rlll}
        \boldsymbol{p}_t & \quad : \quad & \lbrace 1, \ldots , V \rbrace \rightarrow [0,1]\quad & ; \; \boldsymbol{Y}_t \sim  \boldsymbol{p}_t  \label{eqn-step-prob} \\
        \boldsymbol{p}_t(\boldsymbol{y}_t) & \quad \coloneqq \quad &  \mathbb{P}(\boldsymbol{Y}_t{=}\boldsymbol{y}_t | \boldsymbol{y}_{<t}, \boldsymbol{I}) & \label{eqn-step-cond-prob} \\
		\mathbb{P}(\boldsymbol{y}|\boldsymbol{I}) & \quad = \quad & \prod_{t=1}^{\tau} \boldsymbol{p}_t(\boldsymbol{y}_t) & \label{eqn-sequence-prob}
\end{IEEEeqnarray}

As is typical with sequence generators, the training objective here is to maximize probability of the target sequence $\boldsymbol{y}^{\scriptscriptstyle GT}$. We use the standard per-word cross-entropy objective (\autoref{eqn-sequence-obj}), modified slightly for the mini-batch (\autoref{eqn-batch-obj}). We did not use any regularization objective, relying instead on dropout, data-augmentations and synthetic data to provide regularization.
\begin{IEEEeqnarray}{rCll}
        \mathcal{L}_{\scriptscriptstyle seq} \quad & = & \quad -\frac{1}{\tau} \sum_{t} \ln \big( \boldsymbol{p}_t(\boldsymbol{y}_t^{\scriptscriptstyle GT})\big) \quad & ; \; \tau \equiv \text{\small sequence length} \label{eqn-sequence-obj} \\
        \mathcal{L}_{\scriptscriptstyle batch} \quad & = & \quad -\frac{1}{n} \sum_{\scriptscriptstyle batch} \sum_{t} \ln \big( \boldsymbol{p}_t(\boldsymbol{y}_t^{\scriptscriptstyle GT})\big) \quad & ; \; n \equiv \text{\small \# of tokens in batch} \label{eqn-batch-obj}
\end{IEEEeqnarray}

The final Linear layer of the decoder (Figure. \ref{fig-model-outline}) is a 1x1 convolution function that produces logits which are then normalized by softmax to produce $\boldsymbol{p}_t$.

\subsubsection{Combination of Vision and NLP}
One of the strengths of our architecture is in the combination of Vision and Language models.
{\small CNN}s such as ResNet are considered best for processing image data.
And Transformers are considered best for Language Modeling ({\small LM}) and Natural Language Understanding ({\small NLU}) tasks \citep{raffel2020exploring,Radford2018ImprovingLU,devlin2019bert}, posessing properties that are very useful in dealing with noisy and incomplete text that often occurs in real handwriting.
Having both the visual feature map and a language model, the model can do a much better job than one relying on visual features alone.

\subsubsection{Inference}
We use simple greedy decoding, which picks the highest probability token at each step.
Beam search decoding \citep{Graves2008SupervisedSL} did not yield any accuracy improvement indicating that the model is quite opinionated / confident.

\section{Training Configuration and Procedure}
The base configuration uses grayscale images scaled down to 140-150 dots per inch.
Higher resolutions yielded slightly better accuracy at the cost of compute and memory.
We use the 34-layer configuration of ResNet, but have also successfully trained the 18-layer and 50-layer configurations; larger models tending to do better in general as expected.

The following is the base configuration of the Transformer stack:
\begin{itemize}
    \item $N$ (number of layers) = 6
    \item $d_{model}$ = 260
    \item $h$ (number of heads) = 4
    \item $d_{ff}$ (inner-layer of positionwise feed-forward network) = 1024
    \item Activation function inside feed-forward layer = {\small GELU} \citep{DBLP:journals/corr/HendrycksG16}
    \item dropout = 0.5
\end{itemize}

The model was implemented in PyTorch \cite{NEURIPS2019_9015}, and training was carried out using 8 {\small NVIDIA} 2080Ti GPUs.
% Since image-sizes vary significantly between full-page and single-line datasets, different mini-batch sizes were used.
For full page datasets a mini-batch size of 56 combined with a gradient accumulation factor of 2 was used, yielding an effective batch-size of 112.
Single-line datasets had batch sizes as high as 200, but were adjusted downwards when using higher angles of image rotation.
{\small ADAM} optimizer \citep{kingma2017adam} was employed with a fixed learning rate ($\alpha$) of 0.0002, $\beta_{1}$ = 0.9 and $\beta_{2}$ = 0.999.

While all images in a batch must have the same size; we also set all batches to have the same image size, padding smaller images as needed.
This helps during training because any impending {\small GPU OOM} errors surface quickly at the beginning of the run.
It also makes the validation / test results agnostic of the batching scheme since the images will always be the same size regardless of how they are grouped.
Smaller images within a batch are centered during validation and testing.
Padding color can be either the max of 4 corner pixels or simply 0 (black), the choice having no impact on model accuracy.
% We used the first scheme in synthetic data experiments, and the second scheme in Free Form Answers experiments, and
% Results on the Free Form Answers data (\autoref{tab-free-response}) though, were obtained with a fixed padding color of 0 (black).
% These choices did not impact task accuracy.

The base configuration vocabulary consists of all lowercase ASCII printable characters, including space and newline.
% Ground truth labels are programmatically mapped into the vocabulary set.

We observed that model performance increases with increased layer sizes and also image resolution.
It also tends to improve monotonically with training time.
Therefore, we trained our models for as long as possible the longest being 11 days on full pages (\tilda47M samples) and 8 days (\tilda102M samples) on single lines.
Typical training length though is roughly 24M total training samples.
Model state is saved periodically during training.
At the end of training, the checkpoint that performs best on validation is selected for final testing.

\section{Data}\label{sec-data}
\begin{figure}
     \centering
    \subfloat{
        \includegraphics[width=0.45\textwidth]{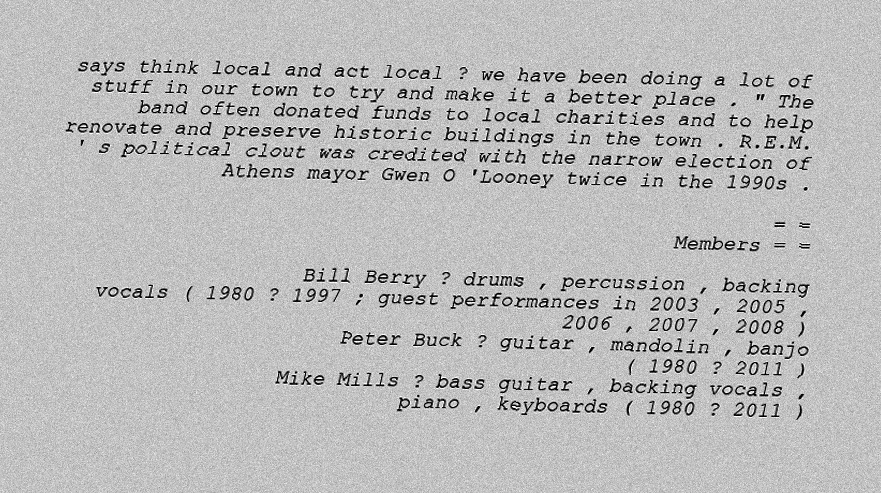}
        \label{fig-wikitext-1col}
    }
    \hfill
    \subfloat{
        \includegraphics[width=0.45\textwidth]{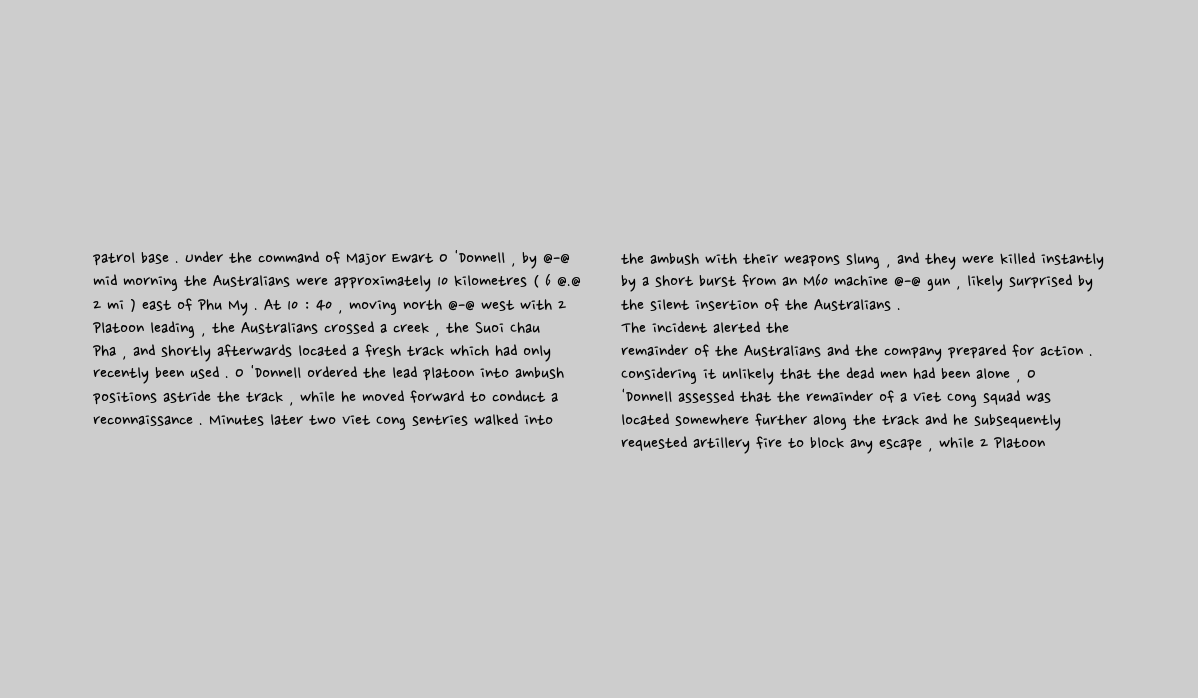}
        \label{fig-wikitext-2col}
    }
    \caption{\small Synthetic samples generated from WikiText-103 Dataset \citep{DBLP:journals/corr/MerityXBS16}. For the two column sample (b), the model -- as per training -- predicts the left column first, then a column separator tag \textless col\textgreater \, and then the right column.}
    \label{fig-synth}
\end{figure}

\subsubsection{Data Sources}
The following is a comprehensive set of all data sources, although each experiment used only a subset:
\begin{itemize}
    \item \textsc{IAM}: The {\small IAM} dataset \citep{IAM-database} with the {\small RWTH} Aachen University split \citep{SLR56}, which is a widely used benchmark for handwriting recognition.
    \item \textsc{WikiText}: The WikiText-103 \citep{DBLP:journals/corr/MerityXBS16} dataset was used to generate synthetic images (explained later).
    \item \textsc{Free Form Answers}: Our proprietary dataset with about 13K samples described in \autoref{sec-free-form-dataset}.
    \item \textsc{Answers2}: Another proprietary dataset of about 16K segmented words and lines extracted from handwritten test answers. However since we train full pages, we stitched back neighboring words and lines using heuristics.
    \item \textsc{Names}: Yet another proprietary dataset of about 22K handwritten names and identifiers.
\end{itemize}
Proprietary datasets were only used for results reported in \autoref{tab-free-response}.

\subsubsection{Synthetic Data Augmentation}\label{sec-synth}
Since the IAM data set has word-level segmentation, we generate synthetic IAM samples by stitching together images of random spans of words or lines. This made it possible to significantly augment the IAM paragraph dataset beyond the mere 747 training forms available in the RWTH Aachen split. Without this augmentation the model would not generalize at all.

We also generate synthetic images on the fly from the WikiText data by picking random text spans and rendering them into single-column and/or two-column layout, using 34 fonts in multiple sizes for a total of 114 combinations (\autoref{fig-synth}). The WikiText data is over 530 million characters long from which over 583 billion unique strings of lengths ranging from 1 to 1100 may be created. Multiplying this with 114 (fonts) yields an epoch size of 66.46 trillion which we would never get through in any of our runs. The dataset thus provides us with a continuous stream of unique samples which builds the implicit language model. Furthermore this trick can be used to `teach' the model new language and terminology by using an appropriate seed text file. Addition of this dataset reduces the error rate on IAM paragraphs by only about 0.4\% (Table \ref{tab-IAM}.) on the IAM dataset but significantly improves the validation loss - about 30\%.

Additionally, we generate empty images of varying backgrounds on the fly. Without this data, the model generates text even on blank images -- which provides evidence of an underlying language model working a little overzealously.

\subsubsection{Image Augmentation}
The following transformations were applied to individual training images: 1) Image scale, 2) rotation, 3) brightness, 4) background color of synthetic images, 5) contrast, 6) perspective and 7) Gaussian noise. At the batch level the images were randomly placed within the larger batch image size during training but centered during validation and testing.

\subsubsection{Data Sampling}
We found that the model generalized best when prior biases were eliminated from the training data. Therefore the datasets are sampled on the fly for every mini-batch with a configurable distribution. We also do not group images by size or sequence length rather we sample them uniformly. Further, parameters for synthetic data generation (e.g. the text span to render or the font to render it in) and image augmentation (e.g. scale of image, angle of rotation, background color) are also sampled uniformly.

Inspite of this sampling scheme, one bias does enter into the training data: padding around text. This is so because most images of a batch are smaller than the batch image size and therefore get padded. This causes the model to perform poorly in inference unless the image was sufficient padding. The optimal padding amount becomes a new hyperparameter that requires tuning after training. We circumvented this problem by padding all images to the same (large) size both during training and inference. This scheme though has the downside of consuming the highest amount of encoder compute regardless of the original image size. Therefore the first scheme is preferable for production.

\section{Results}
Table \ref{tab-IAM} shows character error rates (Levenstein edit distance divided by length of ground truth) of paragraph and single line level tests on the IAM dataset. FPHR refers to our model trained with only the IAM dataset whereas FPHR+Aug refers to our model trained with the IAM plus WikiText based synthetic dataset\footnote{We view synthetic WikiText based data as an augmentation method since it does not rely on proprietary data or method.}.
FPHR trained at \tilda 145 DPI outperforms previous state of art in the paragraph level test\footnote{Results from \citep{8270042} are not included because it was trained on a lot more than IAM  data and 30\% of it was proprietary.} and FPHR+Aug improves error rate by a further 0.4\%.
% Similarly, FPHR trained with 240 dpi images improves the state of art at 300 dpi by about 1\%.
When trained on single lines only, performance is similar to full page but short of the specialized state-of-the-art single-line models. This is because the model does equally well on different text lengths and number of lines provided they were uniformly distributed during training i.e., performance tracks the training data distribution. Notice that our corpus level CERs are lower than the averages indicating better performance on longer text. This is not a characteristic of the model, rather that of the training data which had fewer short sequences. We believe that the observed performance (6+\% test and 4+\% validation CER) is the upper limit of this system and task configuration i.e., the model architecture, its size, the dataset and its split. Increasing the model size and / or image resolution does improve performance slightly (albiet by less than 1\%) as does increasing the amount of training data.

When trained on all datasets and evaluated on Free Form Answers, our model gets a error rate of 7.6\% vs the best available cloud API's 14.4\% (Table \ref{tab-free-response}). Favoring the cloud models, we removed auxiliary markup, line indentations and lower cased all predicted and ground truth text for this comparison\footnote{We evaluated Microsoft, Google and Mathpix cloud APIs. Microsoft performed the best and its results are reported here. This is not intended to be a comparison of models, rather a practical data point that can be used to make build-vs-buy decisions.}.

The model has no trouble transcribing two column layout; its performance on two-column and single-column data are comparable (\autoref{tab-free-response}) providing evidence of its adaptability to different text layouts. The Cloud API on the other hand does well on single-column data, but falters with two-column text precisely because it concatenates adjacent lines across columns.

Inferencing takes an average of 4.6 seconds on a single CPU thread for a set of images averaging 2500x2200 pixels, 456 chars and 11.65 lines without model compression i.e., model pruning, distillation or quantization.

\defcitealias{DBLP:journals/corr/BlucheLM16}{B1}
\defcitealias{Bluche2016JointLS}{B2}
\defcitealias{wang2019decoupled}{W-2019}
\defcitealias{kang2020pay}{K-2020}
\defcitealias{8269951}{P-2017}

\begin{table}
    \centering
    \caption{Comparison on the IAM dataset with and without closed lexicon decoding (LM). Figures in brackets are corpus level scores. $^\bigstar$Model requires paragraph segmentation. $^\ddagger$FPHR trained with single lines only. SLS = Shredding Line Segmentation.}
    \vskip 0.15in
    \begin{small}
    % \begin{sc}
    \begin{tabular}[t]{|l|l|c|c|}
    \toprule
    
    \multirow{2}{*}{ Test Type } & \multirow{2}{*}{ Model } & Mean Test & Mean Test\\
    & & CER w/o LM & CER w/ LM \\
    \midrule
    \multirow{3}{*}{
        \begin{tabular}{l}
            Paragraph\\
            Level
        \end{tabular}
    } & \citet{DBLP:journals/corr/BlucheLM16} (150 dpi) \quad & 16.2\% & \\
    & FPHR (\tilda 145 dpi) & \textbf{6.7} (6.5)\% & \\
    & FPHR+Aug (\tilda 145 dpi) \quad & \textbf{6.3} (6.1)\% & \\
    % FPHR-512+Aug (\tilda 150 dpi) & \textbf{6.8\%} & \textbf{4.2\%} \\
    % \midrule
    % FPHR (240 dpi)\textsuperscript{*} & \textbf{6.9} (6.8)\% \\ %& \textbf{4.8}(4.7)\% \\
    \midrule
    \multirow{4}{*}{
        \begin{tabular}{l}
            Paragraph\\
            Level$^\bigstar$
        \end{tabular}
    } & \citet{Bluche2016JointLS} (150 dpi) & 10.1\% & 6.5\% \\
    & \citet{Bluche2016JointLS} (300 dpi) & 7.9\% & \textbf{5.5}\% \\
    & SLS + MDLSTM + CTC \citep{DBLP:journals/corr/BlucheLM16} (150 dpi) \quad & 11.1\% & \\
    & SLS + MDLSTM + CTC \citep{DBLP:journals/corr/BlucheLM16} (300 dpi) \quad & 7.5\% & \\
    \midrule
    \multirow{5}{*}{ Single Line }
    & \citet{8269951}& 5.8\% & \textbf{4.4}\% \\
    & \citet{Wigington_2018_ECCV}& 6.4\% & \\
    & \citet{wang2019decoupled} & 6.4\% & \\
    & \citet{kang2020pay}& \textbf{4.7}\% &\\
    & FPHR+Aug$^\ddagger$ (\tilda 145 dpi) & 6.5 (5.9)\% & \\
    \bottomrule
    \end{tabular}
    % \end{sc}
    \end{small}
    \label{tab-IAM}
\end{table}

\begin{table}
\caption{Character Error Rates on Free Form Answers and multi column synthetic datasets. $^\star$FPHR trained on one and two col. WikiText synthetic data.}
\label{tab-free-response}
\vskip 0.15in
\begin{center}
\begin{small}
% \begin{sc}
\begin{tabular}{|l|c|r|}
\toprule
Test Data set & Best Cloud API & FPHR \\
\midrule
Free Form Answers    & 14.4\% & \textbf{7.6}\% \\
Wikitext (1 column)   & 1.4\% & 0.008\%$^\star$ \\
Wikitext (2 column)   & 57\% & 0.012\%$^\star$ \\
% Wikitext (0-360\degree orientation) & - & - & TBD \\
\bottomrule
\end{tabular}
% \end{sc}
\end{small}
\end{center}
\vskip -0.1in
\end{table}

\section{Conclusion}
We have presented a ``modern'' neural network architecture that can be trained to perform full page handwriting recognition without image segmentation, delivers state of art accuracy and which is also small and fast enough to be deployed into commercial production. It adapts reasonably to different vocabularies, text layouts and auxiliary tasks and is therefore fit to serve a variety of handwriting and printed text recognition scenarios. The model is also quite easy to replicate using open source `off the shelf' modules making it all the more compelling. Although the overall Full Page HTR problem is not solved yet, we believe it takes us one step forward from \citep{DBLP:journals/corr/BlucheLM16} and \citep{Bluche2016JointLS}.

\subsection{Limitations and Future Work}
Although the presented framework encompasses multiple tasks, available datasets are usually heavily biased towards one or two thereby masking the model's performance on outlier tasks. For e.g., there's usually only one transcribed text region per sample in the Free Form dataset which makes the model tend to transcribe only one (main) text region while skipping others. On the other hand when the dataset is balanced e.g., with one and two column synthetic text, it performs well on both layouts. That said, this aspect needs to be explored more thoroughly and hopefully with standardized datasets and tasks so that the research community can iterate over it.
% Based on current state of art it seems doing so would require a lot more and balanced data and/or a better model architecture. Synthetic real-world data augmentation (\autoref{sec-synth}) and unsupervised pretraining can help alleviate the data size issue to an extent, but regardless this is a non trivial problem.

Second, we have only trained with text up to 1100 characters long and averaging 360 characters. Should there be a need to transcribe longer lengths of text say 10K characters, then some more work becomes necessary in order to deal with longer sequence lengths. Possible solutions include the use of multicharacter vocabularies and sparse Transformers such as \citep{beltagy2020longformer,dai2019transformerxl}.

Other desirable improvements to the model include 1) reducing its sensitivity to image padding. 2) Reducing the encoder's size, which currently stands at almost 22 million parameters. 3) Separating vision models for significantly different visual data (e.g., Synthetic v/s Free Response Answers) so that they may both contribute to the language model but not interfere with each other's vision models.

We believe that the Full Page HTR problem cannot be considered ``solved'' until the error rate has been brought down to less than 1\%. Therefore, more work remains for the community.

\subsubsection{Acknowledgements} We would like to thank Saurabh Bipin Chandra \;for implementing the fast inference path ($O(N^2)$) of the  Transformer decoder, which was lacking in PyTorch.
% \bibliography{fphr}

\begin{thebibliography}{33}
\providecommand{\natexlab}[1]{#1}
\providecommand{\url}[1]{\texttt{#1}}
\expandafter\ifx\csname urlstyle\endcsname\relax
  \providecommand{\doi}[1]{doi: #1}\else
  \providecommand{\doi}{doi: \begingroup \urlstyle{rm}\Url}\fi

\bibitem[Beltagy et~al.(2020)Beltagy, Peters, and Cohan]{beltagy2020longformer}
Beltagy, I., Peters, M.E., and Cohan, A.
\newblock Longformer: The long-document transformer, 2020.

\bibitem[{Bluche} and {Messina}(2017)]{8270042}
{Bluche}, T. and {Messina}, R.
\newblock Gated convolutional recurrent neural networks for multilingual
  handwriting recognition.
\newblock In \emph{2017 14th IAPR International Conference on Document Analysis
  and Recognition (ICDAR)}, volume~01, pages 646--651, 2017.

\bibitem[Bluche(2016)]{Bluche2016JointLS}
Bluche, T.
\newblock Joint line segmentation and transcription for end-to-end handwritten
  paragraph recognition.
\newblock \emph{ArXiv}, abs/1604.08352, 2016.

\bibitem[Bluche et~al.(2016)Bluche, Louradour, and
  Messina]{DBLP:journals/corr/BlucheLM16}
Bluche, T., Louradour, J., and Messina, R.O.
\newblock Scan, attend and read: End-to-end handwritten paragraph recognition
  with {MDLSTM} attention.
\newblock \emph{CoRR}, abs/1604.03286, 2016.

\bibitem[Dai et~al.(2019)Dai, Yang, Yang, Carbonell, Le, and
  Salakhutdinov]{dai2019transformerxl}
Dai, Z., Yang, Z., Yang, Y., Carbonell, J., Le, Q.V., and Salakhutdinov, R.
\newblock Transformer-xl: Attentive language models beyond a fixed-length
  context, 2019.

\bibitem[Deng et~al.(2017)Deng, Kanervisto, Ling, and
  Rush]{Deng2017ImagetoMarkupGW}
Deng, Y., Kanervisto, A., Ling, J., and Rush, A.M.
\newblock Image-to-markup generation with coarse-to-fine attention.
\newblock In \emph{ICML}, 2017.

\bibitem[Devlin et~al.(2019)Devlin, Chang, Lee, and Toutanova]{devlin2019bert}
Devlin, J., Chang, M.W., Lee, K., and Toutanova, K.
\newblock Bert: Pre-training of deep bidirectional transformers for language
  understanding, 2019.

\bibitem[Graves(2008)]{Graves2008SupervisedSL}
Graves, A.
\newblock Supervised sequence labelling with recurrent neural networks.
\newblock In \emph{Studies in Computational Intelligence}, 2008.

\bibitem[Graves et~al.(2006)Graves, Fern{\'a}ndez, Gomez, and
  Schmidhuber]{Graves2006ConnectionistTC}
Graves, A., Fern{\'a}ndez, S., Gomez, F., and Schmidhuber, J.
\newblock Connectionist temporal classification: labelling unsegmented sequence
  data with recurrent neural networks.
\newblock In \emph{ICML '06}, 2006.

\bibitem[Graves and Schmidhuber(2008)]{Graves2008OfflineHR}
Graves, A. and Schmidhuber, J.
\newblock Offline handwriting recognition with multidimensional recurrent
  neural networks.
\newblock In \emph{NIPS}, 2008.

\bibitem[He et~al.(2015)He, Zhang, Ren, and Sun]{DBLP:journals/corr/HeZRS15}
He, K., Zhang, X., Ren, S., and Sun, J.
\newblock Deep residual learning for image recognition.
\newblock \emph{CoRR}, abs/1512.03385, 2015.

\bibitem[Hendrycks and Gimpel(2016)]{DBLP:journals/corr/HendrycksG16}
Hendrycks, D. and Gimpel, K.
\newblock Bridging nonlinearities and stochastic regularizers with gaussian
  error linear units.
\newblock \emph{CoRR}, abs/1606.08415, 2016.
\newblock URL \url{http://arxiv.org/abs/1606.08415}.

\bibitem[Kang et~al.(2020)Kang, Riba, Rusiñol, Fornés, and
  Villegas]{kang2020pay}
Kang, L., Riba, P., Rusiñol, M., Fornés, A., and Villegas, M.
\newblock Pay attention to what you read: Non-recurrent handwritten text-line
  recognition, 2020.

\bibitem[Kingma and Ba(2017)]{kingma2017adam}
Kingma, D.P. and Ba, J.
\newblock Adam: A method for stochastic optimization, 2017.

\bibitem[{Lecun} et~al.(1998){Lecun}, {Bottou}, {Bengio}, and
  {Haffner}]{726791}
{Lecun}, Y., {Bottou}, L., {Bengio}, Y., and {Haffner}, P.
\newblock Gradient-based learning applied to document recognition.
\newblock \emph{Proceedings of the IEEE}, 86\penalty0 (11):\penalty0
  2278--2324, 1998.

\bibitem[Ly et~al.(2019)Ly, Nguyen, and Nakagawa]{8978049}
Ly, N.T., Nguyen, C.T., and Nakagawa, M.
\newblock An attention-based end-to-end model for multiple text lines
  recognition in japanese historical documents.
\newblock In \emph{2019 International Conference on Document Analysis and
  Recognition (ICDAR)}, pages 629--634, 2019.
\newblock \doi{10.1109/ICDAR.2019.00106}.

\bibitem[Marti and Bunke(2002)]{IAM-database}
Marti, U.V. and Bunke, H.
\newblock The iam-database: An english sentence database for offline
  handwriting recognition.
\newblock \emph{International Journal on Document Analysis and Recognition},
  5:\penalty0 39--46, 11 2002.
\newblock \doi{10.1007/s100320200071}.

\bibitem[Merity et~al.(2016)Merity, Xiong, Bradbury, and
  Socher]{DBLP:journals/corr/MerityXBS16}
Merity, S., Xiong, C., Bradbury, J., and Socher, R.
\newblock Pointer sentinel mixture models.
\newblock \emph{CoRR}, abs/1609.07843, 2016.

\bibitem[{Open SLR}()]{SLR56}
{Open SLR}.
\newblock Aachen data splits (train, test, val) for the iam dataset.
\newblock \url{https://www.openslr.org/56/}.
\newblock Identifier: SLR56.

\bibitem[Parmar et~al.(2018)Parmar, Vaswani, Uszkoreit, Łukasz Kaiser,
  Shazeer, Ku, and Tran]{parmar2018image}
Parmar, N., Vaswani, A., Uszkoreit, J., Łukasz Kaiser, Shazeer, N., Ku, A.,
  and Tran, D.
\newblock Image transformer, 2018.

\bibitem[Paszke et~al.(2019)Paszke, Gross, Massa, Lerer, Bradbury, Chanan,
  Killeen, Lin, Gimelshein, Antiga, Desmaison, Kopf, Yang, DeVito, Raison,
  Tejani, Chilamkurthy, Steiner, Fang, Bai, and Chintala]{NEURIPS2019_9015}
Paszke, A., Gross, S., Massa, F., Lerer, A., Bradbury, J., Chanan, G., Killeen,
  T., Lin, Z., Gimelshein, N., Antiga, L., Desmaison, A., Kopf, A., Yang, E.,
  DeVito, Z., Raison, M., Tejani, A., Chilamkurthy, S., Steiner, B., Fang, L.,
  Bai, J., and Chintala, S.
\newblock Pytorch: An imperative style, high-performance deep learning library.
\newblock In Wallach, H., Larochelle, H., Beygelzimer, A., d\textquotesingle
  Alch\'{e}-Buc, F., Fox, E., and Garnett, R., editors, \emph{Advances in
  Neural Information Processing Systems 32}, pages 8024--8035. Curran
  Associates, Inc., 2019.
\newblock URL
  \url{http://papers.neurips.cc/paper/9015-pytorch-an-imperative-style-high-performance-deep-learning-library.pdf}.

\bibitem[Pham et~al.(2013)Pham, Kermorvant, and
  Louradour]{DBLP:journals/corr/PhamKL13}
Pham, V., Kermorvant, C., and Louradour, J.
\newblock Dropout improves recurrent neural networks for handwriting
  recognition.
\newblock \emph{CoRR}, abs/1312.4569, 2013.
\newblock URL \url{http://arxiv.org/abs/1312.4569}.

\bibitem[{Puigcerver}(2017)]{8269951}
{Puigcerver}, J.
\newblock Are multidimensional recurrent layers really necessary for
  handwritten text recognition?
\newblock In \emph{2017 14th IAPR International Conference on Document Analysis
  and Recognition (ICDAR)}, volume~01, pages 67--72, 2017.

\bibitem[Radford(2018)]{Radford2018ImprovingLU}
Radford, A.
\newblock Improving language understanding by generative pre-training.
\newblock 2018.

\bibitem[Raffel et~al.(2020)Raffel, Shazeer, Roberts, Lee, Narang, Matena,
  Zhou, Li, and Liu]{raffel2020exploring}
Raffel, C., Shazeer, N., Roberts, A., Lee, K., Narang, S., Matena, M., Zhou,
  Y., Li, W., and Liu, P.J.
\newblock Exploring the limits of transfer learning with a unified text-to-text
  transformer, 2020.

\bibitem[Singh(2018)]{DBLP:journals/corr/abs-1802-05415}
Singh, S.S.
\newblock Teaching machines to code: Neural markup generation with visual
  attention.
\newblock \emph{CoRR}, abs/1802.05415, 2018.

\bibitem[Sutskever et~al.(2014)Sutskever, Vinyals, and
  Le]{DBLP:journals/corr/SutskeverVL14}
Sutskever, I., Vinyals, O., and Le, Q.V.
\newblock Sequence to sequence learning with neural networks.
\newblock \emph{CoRR}, abs/1409.3215, 2014.
\newblock URL \url{http://arxiv.org/abs/1409.3215}.

\bibitem[Vaswani et~al.(2017)Vaswani, Shazeer, Parmar, Uszkoreit, Jones, Gomez,
  Kaiser, and Polosukhin]{DBLP:journals/corr/VaswaniSPUJGKP17}
Vaswani, A., Shazeer, N., Parmar, N., Uszkoreit, J., Jones, L., Gomez, A.N.,
  Kaiser, L., and Polosukhin, I.
\newblock Attention is all you need.
\newblock \emph{CoRR}, abs/1706.03762, 2017.

\bibitem[Vaswani et~al.(2018)Vaswani, Bengio, Brevdo, Chollet, Gomez, Gouws,
  Jones, Kaiser, Kalchbrenner, Parmar, Sepassi, Shazeer, and
  Uszkoreit]{DBLP:journals/corr/abs-1803-07416}
Vaswani, A., Bengio, S., Brevdo, E., Chollet, F., Gomez, A.N., Gouws, S.,
  Jones, L., Kaiser, L., Kalchbrenner, N., Parmar, N., Sepassi, R., Shazeer,
  N., and Uszkoreit, J.
\newblock Tensor2tensor for neural machine translation.
\newblock \emph{CoRR}, abs/1803.07416, 2018.

\bibitem[{Voigtlaender} et~al.(2016){Voigtlaender}, {Doetsch}, and
  {Ney}]{7814068}
{Voigtlaender}, P., {Doetsch}, P., and {Ney}, H.
\newblock Handwriting recognition with large multidimensional long short-term
  memory recurrent neural networks.
\newblock In \emph{2016 15th International Conference on Frontiers in
  Handwriting Recognition (ICFHR)}, pages 228--233, 2016.

\bibitem[Wang et~al.(2019)Wang, Zhu, Jin, Luo, Chen, Wu, Wang, and
  Cai]{wang2019decoupled}
Wang, T., Zhu, Y., Jin, L., Luo, C., Chen, X., Wu, Y., Wang, Q., and Cai, M.
\newblock Decoupled attention network for text recognition, 2019.

\bibitem[Wigington et~al.(2018)Wigington, Tensmeyer, Davis, Barrett, Price, and
  Cohen]{Wigington_2018_ECCV}
Wigington, C., Tensmeyer, C., Davis, B., Barrett, W., Price, B., and Cohen, S.
\newblock Start, follow, read: End-to-end full-page handwriting recognition.
\newblock In \emph{Proceedings of the European Conference on Computer Vision
  (ECCV)}, September 2018.

\bibitem[Xu et~al.(2015)Xu, Ba, Kiros, Cho, Courville, Salakhutdinov, Zemel,
  and Bengio]{Xu2015ShowAA}
Xu, K., Ba, J., Kiros, J.R., Cho, K., Courville, A.C., Salakhutdinov, R.,
  Zemel, R.S., and Bengio, Y.
\newblock Show, attend and tell: Neural image caption generation with visual
  attention.
\newblock In \emph{ICML}, 2015.

\end{thebibliography}

\end{document}